\documentclass[final]{cvpr}

\usepackage{times}
\usepackage{epsfig}
\usepackage{graphicx}
\usepackage{amsmath}
\usepackage{amssymb}
\usepackage{multirow}
\usepackage{bm}
\usepackage[normalem]{ulem}
\useunder{\uline}{\ul}{}
\usepackage{siunitx}
\usepackage{array,booktabs,makecell}
    
\usepackage{subcaption}
\usepackage[dvipsnames]{xcolor}

\usepackage[ruled,vlined,linesnumbered]{algorithm2e}

\usepackage[pagebackref=false,breaklinks=true,colorlinks,bookmarks=false]{hyperref}

\def\cvprPaperID{5} % *** Enter the CVPR Paper ID here
\def\confYear{BAMFG 2021}

\begin{document}
\pagenumbering{gobble}
%%%%%%%%% TITLE
\title{Part-aware Measurement for Robust Multi-View Multi-Human 3D Pose Estimation and Tracking}
\author{Hau Chu$^{1,3}$ \qquad Jia-Hong Lee$^{1,3}$ \qquad Yao-Chih Lee$^{2,5}$ \qquad Ching-Hsien Hsu$^{2,6}$ \\Jia-Da Li$^{1,3}$ \qquad Chu-Song Chen$^{1,4}$ \\
$^1$National Taiwan University \qquad $^2$Academia Sinica, Taiwan \\
\tt\small $^3$\{hauchu8733,jiahonghenrylee,jiadali\}@ntu.edu.tw, $^4$chusong@csie.ntu.edu.tw \\
\tt\small $^5$yclee1231@iis.sinica.edu.tw, $^6$applychristw@gmail.com}

\maketitle

%%%%%%%%% ABSTRACT
\begin{abstract}
% original abstract
This paper introduces an approach for multi-human 3D pose estimation and tracking based on calibrated multi-view. The main challenge lies in finding the cross-view and temporal correspondences correctly even when several human pose estimations are noisy. Compare to previous solutions that construct 3D poses from multiple views, our approach takes advantage of temporal consistency to match the 2D poses estimated with previously constructed 3D skeletons in every view. Therefore cross-view and temporal associations are accomplished simultaneously. Since the performance suffers from mistaken association and noisy predictions, we design two strategies for aiming better correspondences and 3D reconstruction.  Specifically, we propose a part-aware measurement for 2D-3D association and a filter that can cope with 2D outliers during reconstruction. Our approach is efficient and effective comparing to state-of-the-art methods; it achieves competitive results on two benchmarks: 96.8\% on Campus and 97.4\% on Shelf. Moreover, we extends the length of Campus evaluation frames to be more challenging and our proposal also reach well-performed result. The code will be available at \url{https://git.io/JO4KE}.
\end{abstract}

%%%%%%%%% BODY TEXT
\begin{figure*}[t]
    \begin{center}
      \includegraphics[width=0.96\linewidth]{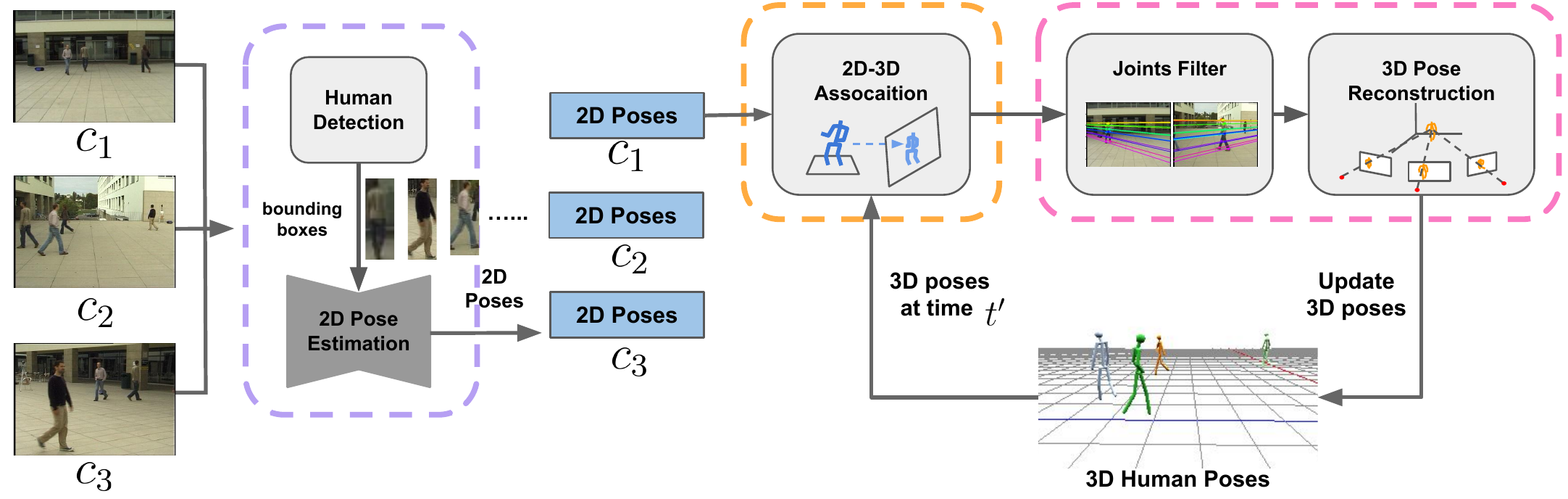}
   \end{center}
   \caption{Overview of our 3D pose tracking mechanism. First, initial 3D poses are given via the reconstruction with cross-view 2D correspondences. Then, given new frames as shown in the figure, we apply 2D-3D association to match estimated 2D poses with preciously tracked 3D skeletons. While finishing the association of all views, for each matched 2D poses and 3D skeleton, \textit{joints filter} is employed to remove 2D outliers of estimated body joints and update the 3D pose.}
   \vspace{-0.2cm} 
   \label{fig:overview}
\end{figure*}
\section{Introduction}
Multi-human 3D pose estimation and tracking based on multi-view streaming videos have many applications, including marker-less motion capture~\cite{liu2013markerless,elhayek2016marconi,tome2018rethinking,pavlakos2017harvesting}, sports analysis~\cite{Bridgeman_2019_CVPR_Workshops}, and video surveillance~\cite{chen2020cross}. Recently, plenty of nicely performed 2D pose estimation approaches~\cite{chen2018cascaded,HRNet,cao2019openpose} have been developed; they have then been extended to the estimation of 3D poses from a monocular view~\cite{guler2018densepose,jiang2020coherent}. Although great progress has been made on inferring the 3D human poses in a single view, the noisy predictions caused by large pose variations and partial occlusions remain to be demanding. To problems, address these issues, constructing 3D human poses from multi-camera views becomes a promising way. However, there are still several challenges to be tackled, such as heavy occlusions, high computational complexity, and noisy predictions. 

In general, given synced multi-view video input, three main tasks should be dealing with properly: \textit{2D skeleton extraction}, \textit{cross-view association}, and \textit{temporal association}. \noindent Human skeletons are estimated for each view initially, which is often achieved via a 2D human pose estimation method with convolution neural networks (CNNs)~\cite{chen2018cascaded,Hourglass,HRNet}. \noindent 
The 3D skeletons are then reconstructed according to the associated 2D human skeletons in different views. \noindent In the last, associations between the 2D or 3D skeletons are established with those in the next frame in video streams.

Leveraging 2D poses, recent studies~\cite{Bridgeman_2019_CVPR_Workshops,tanke2019iterative,dong2019fast,chen2020cross} follow an initialization-and-tracking framework for 3D pose inference. The framework assumes a streaming mode, i.e., the outputs of multi-view 3D poses are obtained on-line per input frame, and the previously generated outputs cannot be altered. To perform on-line 3D pose inference, initial 3D human skeletons are computed given the first views. This is often achieved via epi-polar-line distance and/or person re-id matching; then, the initial 3D joints are established with multi-view stereo. As for maintaining the flexibility of use, most works~\cite{tanke2019iterative,Bridgeman_2019_CVPR_Workshops,chen2020cross,20204DAssociation} do not need fine-tuning on the data collected in the testing environment; pre-trained person re-id models are thus not effective enough due to the testing domain shift in the usual. Hence, epi-polar-line distances are mostly used for initializing the 3D poses. Once initialized, the obtained 3D skeletons of the individuals serve as a guide for human tracking based on the 2D poses of all views detected in future frames. The 3D human skeletons are then updated via multi-view reconstruction according to the new correspondences of the 2D human skeletons obtained when tracking is finished.

In the processing flow of on-line 3D pose inference, temporal association plays a main role. Previous approaches such as~\cite{dong2019fast,tanke2019iterative,voxelpose} construct the 3D skeletons from multiple current views at first, and then the 3D skeletons obtained are smoothed temporally for each individual. However, since matching of $C$ views involves $O(C^2)$ pairs of the epi-polar-line distance evaluation, the computation cost could be considerably increased with the number of people and cameras. To alleviate the computation burden, we introduce a 2D-3D matching mechanism to enforce the temporal consistency, which leverages the previously tracked 3D skeletons for finding the correspondences among views. The complexity is $O(C)$ as only the projections of the previous 3D skeletons on the $C$ views is needed. Besides, multi-person interactions in crowded scenes would increase the difficulty of matching and identifying people as individuals across views. To exploit the projected 3D skeletons for finding better correspondences among views, we propose a part-aware measurement to compute the affinity. We also design a greedy algorithm called \textit{joints filter} to remove noisy points and construct a robust 3D pose by taking advantage of epi-polar constraints. Unlike previous works such as~\cite{chen2020cross}, the proposed method can handle wrong keypoint estimation caused by occlusion or motion blur during association. We compare our method with previous works and show that our solution achieves competitive results on two benchmark datasets: Campus and Shelf.

In the following contents, Section~\ref{sec:related_work} reviews the related approaches of 3D human pose estimation and tracking. Section~\ref{sec:method} presents the details of our new approach. Section~\ref{sec:exp} presents the results comparing the accuracy and efficiency between our approaches and competitive methods on the Campus~\cite{belagiannis20153d} and Shelf~\cite{belagiannis20153d} datasets.
We also demonstrate the design analysis of our approach in ablation studies. 
Finally, conclusions are given in Section~\ref{sec:conclusion}.

%------------------------------------------------------------------------
\section{Related Work} \label{sec:related_work}
This section reviews the previous approaches to 3D human pose estimation and their tracking techniques.
In Section~\ref{sec:3d_pose_estimation}, we review the methods of 3D human pose estimation, which infer the 3D poses based on isolated images.
In Section~\ref{3d_pose_tracking}, we review the techniques of 3D human pose tracking, where temporal consistency in 3D space is utilized for performance enhancement.

\subsection{3D Human Pose Estimation}
\label{sec:3d_pose_estimation}
Depending on the number of input cameras, 3D human pose estimation methods are divided into a monocular camera for taking single-view video~\cite{andriluka2010monocular,moreno20173d,sun2018integral,guler2018densepose,mehta2018single,cheng2019occlusion,moon2019camera,jiang2020coherent,zhang2020inference} and multiple cameras for taking multi-view videos synchronously~\cite{belagiannis20153d,ershadi2018multiple,Bridgeman_2019_CVPR_Workshops,tanke2019iterative,dong2019fast,pirinen2019domes,chen2020cross,voxelpose,20204DAssociation,tome2018rethinking}.

Due to the difficulty of multi-person 3D poses reconstruction in monocular view, most of the single-view approaches are developed to construct a single person's 3D poses~\cite{moreno20173d,sun2018integral,cheng2019occlusion,zhang2020inference}, where the predicted pose does not include absolute locations in the environment. Therefore, it will limit these approaches applying in different practical surveillance scenarios. 
% modified
Despite a great achievement of multi-human 3D pose estimation in a single view~\cite{andriluka2010monocular,moon2019camera,guler2018densepose}, there is still a large deviation when applying these techniques in different practical surveillance scenarios. In particular, the motion blur and occlusions occur in images.

%To robustly predict multi-person 3D poses from a monocular view, Andriluka \etal~\cite{andriluka2010monocular} utilize the human detector and tracker to obtain the people's locations one by one in the video before predicting 3D human poses. Like Andriluka \etal~\cite{andriluka2010monocular}, Moon \etal~\cite{moon2019camera} utilize Mask R-CNN~\cite{he2017mask} to crop the human region in the monocular image before predicting 2D keypoints. They developed RootNet to crop depth maps and PoseNet to predict 3D keypoints. By combining information of RootNet, the results of PoseNet are further refined.

%Guler \etal~\cite{guler2018densepose} develop the RCNN-based multi-task human detector to detect human locations and 3D poses with depth-based representation. Compared with Guler \etal~\cite{guler2018densepose}, Jiang \etal~\cite{jiang2020coherent} develop the RCNN-based human detector with SMPL parametric body model~\cite{loper2015smpl} to detect the humans' locations and 3D poses with shape-based representation. Despite a great achievement of multi-human 3D pose estimation in a single view, there still is a large deviation when applying these techniques in different practical surveillance scenarios. In particular, the motion blur and occlusions occur in images.

To retrieve absolute location and handle occlusions, the studies of multi-view 3D pose estimation attract more attention recently. It can be applied in various applications, such as sports analysis, video surveillance, animation, and healthcare. Most previous approaches~\cite{belagiannis20153d,pavlakos2017harvesting,qiu2019cross} for single-person 3D pose estimation are developed based on the 3D Pictorial Structure model (3DPS)~\cite{burenius20133d}, which discretizes the 3D-space as a grid and assigns each joint to one assumed location in the grid. Depending on the cross-view association, 3D pose reconstruction can be solved by minimizing the geometric error~\cite{andrew2001multiple} between assumed 3D poses and the 3D poses generated from multi-view 2D images. Since CNN has a great performance in the human detector and 2D human pose estimation, most multi-person 3D pose estimation approaches~\cite{ershadi2018multiple,dong2019fast,voxelpose} utilize sophisticated CNN-based 2D human pose estimation techniques at the beginning. It detects the human in the image and applies 2D pose estimation in the cropped image to obtain feature maps of present humans in these multi-view images. Feature maps from different views will serve various fusion, matching, or clustering techniques to achieve multi-person 3D poses estimation. 

Ershadi-N. \etal~\cite{ershadi2018multiple} utilize DeeperCut~\cite{insafutdinov2016deepercut} to detect the human body's 2D part poses in each image for constructing the initial 3D joints of human body poses in 3D state space. They develop a clustering algorithm to separate all 3D candidate joints into multiple individual 3D human poses before refining these poses using a fully connected pairwise conditional random field (CRF). Dong \etal~\cite{dong2019fast} utilize Faster R-CNN~\cite{ren2016faster} with lightweight backbone network and Cascaded Pyramid Network~\cite{chen2018cascaded} to detect humans' location and their 2D poses in multi-view images. They develop a multi-way matching method with circle consistency to identify the same person across cameras via human affinity before constructing 3D human poses. The human affinity is calculated from CamStyle~\cite{zhong2018camera} generated features and geometric compatibility. Tu \etal~\cite{voxelpose} utilize HRNet~\cite{HRNet} to obtain 2D pose heatmaps in each camera view. To fuse projected 2D pose heatmaps into 3D space, they develop a Cuboid Proposal Network and Pose Regression Network to localize people and regress cuboid proposals to the detailed 3D poses. Their novel solution performs well on both benchmarks, yet retraining the model is necessary when it comes to new scenes. Although these approaches can better perform 3D human pose estimation, the computational cost of cross-view matching and 3D human pose estimation is still too high for real-time applications.

\subsection{3D Human Pose Tracking}
\label{3d_pose_tracking}
To process synchronized multiple camera streaming videos, tracking plays an important role in practice by leveraging temporal information to smooth 3D motion capture, handle current pose ambiguity, and accelerate the system's processing time. Recently, Taylor \etal~\cite{taylor2010dynamical} combine Conditional Restricted Boltzmann Machine~\cite{taylor2007modeling} with Bayesian filtering for tracking single-person 3D poses. With an increasing number of people in the scene, dealing with frequent occlusions and ambiguities becomes crucial. 
Liu \etal~\cite{liu2013markerless} combine image segmentation with articulated template models for accurately capturing and tracking 3D articulated skeleton motion of each person. Like the purpose of Liu \etal~\cite{liu2013markerless}, Elhayek \etal~\cite{elhayek2016marconi} and Li \etal~\cite{li2018shape} combine the CNN-based joint detection method with a model-based motion or spatio-temporal tracking algorithm in a unified 3D pose optimization for tracking and capturing 3D articulated skeleton motion of each person. 

While in larger indoor/outdoor environments with more people and cameras, most approaches~\cite{Bridgeman_2019_CVPR_Workshops,tanke2019iterative,chen2020cross,20204DAssociation} focus on reducing the computation cost while obtaining better performance. Tanke \etal~\cite{tanke2019iterative} utilize a 2D human pose detector to obtain multiple 2D estimated human poses from multiple views and solve the k-partite matching problem using epipolar geometry to build associations among these multiple 2D estimated human poses across multiple views. They thus construct 3D human pose of each person, followed by a greedy algorithm to match and track iteratively across frames. Like Tanke \etal~\cite{tanke2019iterative}, Bridgeman \etal~\cite{Bridgeman_2019_CVPR_Workshops} applies a greedy fashion to seek the best correspondence between views. Estimated 3D skeletons are also exploited as input to improve the tracking quality. Recently, Chen \etal~\cite{chen2020cross} develop a novel 3D human pose tracking technique that speeds up tracking tasks in their large-scale camera systems. Compare to Tanke's work~\cite{tanke2019iterative}, their cross-view tracking with geometric affinity can track and construct 3D human poses of each person across views and frames synchronously. To attain better performance, Zhang \etal~\cite{20204DAssociation} develop a novel 4D (2D spatial, 1D viewpoint, and 1D temporal) association graph algorithm to enhance the accuracy of finding associations among 2D and 3D human poses from views and frames. Our approach is developed by referring to the inspiration of these approaches~\cite{chen2020cross,tanke2019iterative}.
%------------------------------------------------------------------------
\begin{figure*}[t]
   \centering
   \includegraphics[width=.82\linewidth]{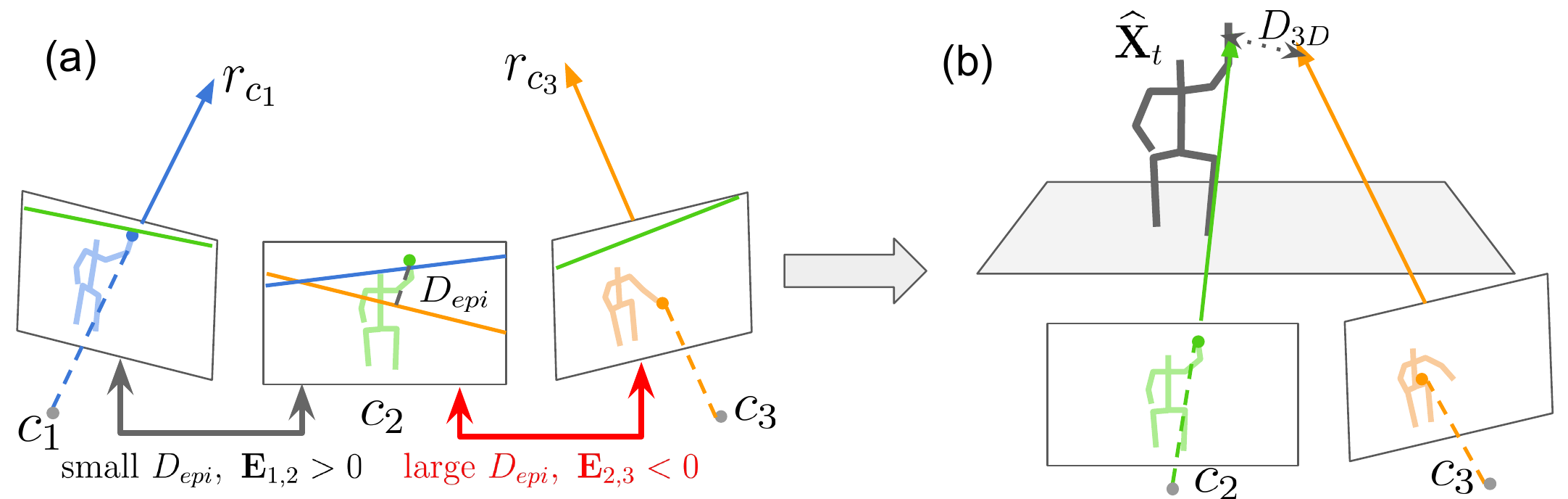}
   \caption{Illustration of \textit{joints filter} algorithm. (a) The blue and orange lines on image plane of $c_2$ are epipolar lines of back-project-ray $r_{c_1}$, $r_{c_3}$, respectively. The joints in views $c_2$ and $c_3$ are considered as outlier candidates due to the large epipolar distance thus be examined in the next step. (b) We remove the outlier with the larger distance between back-project-ray of 2D joint and estimated 3D skeleton $\widehat{\mathbf{X}}_t$.
   }
   \label{fig:Joints_Filter}
\end{figure*}
\section{Methodology} \label{sec:method}
Given multiple cameras $\mathbf{C}$ from different views in a scene, coupling with their known camera intrinsic $\mathbf{K}$ and extrinsic (rotation $\mathbf{R}$ and translation $\mathbf{o}$) information, we aim to find out an unknown number of humans' 3D locations in an area. The proposed 3D reconstruction and tracking framework (as shown in Figure~\ref{fig:overview}) can be divided in to three stages: 2D human pose estimation, 2D-3D association and 3D human pose reconstruction. The overall procedure is shown in Algorithm~\ref{alg:estimation_and_tracking}.

\subsection{2D Human Pose Estimation}
To reconstruct 3D human pose via multiple views, the potential 2D human poses $\mathbf{p}$ in each view are first extracted via an off-the-shelf 2D human pose estimator. It is worth noting that the error of 2D pose estimation may easily affect the following association and 3D pose reconstruction processes, yielding mis-matching or inaccurate 3D pose.
We adopt YOLOv3~\cite{redmon2017yolo9000} coupled with HRNet~\cite{HRNet} as our top-down 2D pose estimator. 
Since HRNet~\cite{HRNet} conducted multi-scale fusions by fusing high and low resolution representations, leading to a potentially more accurate and spatially more precise heatmap predictions, which is different form earlier works that recovered high-resolution representations from lower-resolution representations~\cite{CPN2018}. This combination perform a more efficient and accurate estimation. Yet the comparisons of these off-the-shelf 2D pose estimators are beyond the scope of this paper.
\subsection{Part-aware 2D-3D Association}
\label{sec:partial_matching}
After extracting 2D human poses $\mathbf{p}$ from each view, each 2D pose $\bm{x}_{t,c} \in \mathbf{p}$ is associated with previously reconstructed 3D poses $\mathbf{X}_{t'} \in \mathbf{P}$, which are initialized across views and updated at $t'$ (described in Section~\ref{sec:3D_Reconstruction}.) The affinity $\mathcal{G}(\bm{x}_{t,c}, \mathbf{X}_{t'})$ of a 2D pose $\bm{x}_{t,c}$ to a 3D pose $\mathbf{X}_{t'}$ is measured with the geometric constraints of projection difference between $\bm{x}_{t,c}$ and re-projecting $\mathbf{X}_{t'}$ onto view $c$. The geometry affinity of the $n^{\text{th}}$ joint $\mathcal{G}^n$ is computed by the equation below:
\begin{equation}
   \label{eq:affinity-2d}
   \mathcal{G}^n(\bm{x}^{n}_{t,c},\mathbf{X}^{n}_{t'}) =
        (1 - \frac{
      \left \| \bm{x}^n_{t,c} - \bm{\tilde{x}}^n_{t',c} \right \|
      }{
      \alpha_{2D} (t - t')
      }) \cdot e^{-\lambda_{a} (t-t')}.
\end{equation}
where $\bm{\tilde{x}}^n_{t',c}$ is the 2D projection of $\mathbf{X}^n_{t'}$ on camera $c$.  $\alpha_{2D}$, $\lambda_{\alpha}$ are constants of 2D velocity threshold and penalty rate of time interval respectively.

Common approaches conduct body-aware affinity measurement to acquire a 2D pose affinity to 3D pose by averaging all the joint errors. However, the potential noisy predictions due to occlusion or motion blur may yield mis-matching results. To reduce the effect of noisy estimation of joints, we propose a part-aware measurement approach that can handle those noisy joints (shown in Figure~\ref{fig:match_in_body_or_parts}). In contrast to directly averaging all joint affinities, we only take joints that have positive affinities into consideration, since the outlier joints may yield negative affinities that influence associations. 

Therefore, the affinity $\mathcal{G}(\bm{x}_{t,c}, \mathbf{X}_{t'})$ is the mean of the joint affinities with positive values, indicating the similarity of $\bm{x}^n_{t,c}$ and $\mathbf{X}^n_{t'}$. And for those number of joints with positive affinities smaller than $\varepsilon$, we set $\mathcal{G}(\bm{x}_{t,c}, \mathbf{X}_{t'})$ to 0.

Once the affinities of all pairs of 2D poses and 3D poses are computed, an affinity matrix $\mathbf{A} \in \mathbb{R}^{|\mathbf{P}| \times |\mathbf{p}|}$ can be formed to associate 2D poses to 3D skeletons. This becomes a weighted bipartite problem and can easily be solved by Hungarian algorithm~\cite{Kuhn55thehungarian}. While finishing the associations of all views, cross-view association is accomplished implicitly, where the corresponding 2D poses across views are those assigned to the same 3D skeleton. 
% ---------------------------------------------------------------------------------------------------

\subsection{3D Pose Reconstruction}
\label{sec:3D_Reconstruction}
As long as a tracked 3D skeleton $\mathbf{X}_{t'}$ is matched with 2D poses after association, the new 3D skeleton can be reconstructed with the associated 2D poses from multiple views. However, the association may sometimes gives only single view of 2D pose thus cannot perform reconstruction from multiple views. 
% Therefore, supposing that there are $\mathbf{M}$ cameras got a 2D pose matched with a 3D skeleton, we then gather a set $\mathbb{P}= \{\bm{x}_{t,c} | c \in \mathbf{M}\} \cup \{\bm{x}_{t'_{c}, c} | c\in \mathbf{C-M}; 0\leq t-t'_{c} < \tau\}$
% modified
Therefore, we gather a set $\mathbb{P}= \{\bm{x}_{t'_{c},c} | c\in \mathbf{C}; 0\leq t-t'_{c} < \tau\}$, where $t'_{c}$ denotes the time of the last matched 2D pose from camera $c$ and $t'_c \leq t$, to retrieve latest associated 2D pose in each camera within a short time interval $\tau$ for triangulation, and we set $\tau$ as 75-100ms empirically in practice. Furthermore, we propose \textit{joints filter} as pre-processing of reconstruction to handle noisy joint estimations.\\
\noindent\textbf{Joints Filter} is also a part-aware processing. In some cases, noisy predictions occur because of partial occlusion or motion blur. Epipolar constraint is utilized to remove outliers by computing the distance between epipolar line and the corresponding point. Here we handle each joint independently. Therefore, an epipolar affinity matrix of the $n^{\text{th}}$ joint $\mathbf{E} \in \mathbb{R}^{|\mathbb{P}| \times |\mathbb{P}|}$ can be computed. 
\begin{equation}
    \label{eq:epipolar_distance}
    \mathbf{E}_{i,j}(\bm{x}^{n}_{i},\bm{x}^{n}_{j}) = 1 - \frac{d(\bm{x}^{n}_{i}, L_{j}) + d(\bm{x}^{n}_{j}, L_{i})}{2 \alpha_{epi}}.
\end{equation}
where $L$, $d$ denote the epipolar line and the point-to-line distance function, respectively. $\alpha_{epi}$ represents a threshold of acceptable distance error. Whenever there is a negative value at $\mathbf{E}_{i,j}$, at least one of $\bm{x}^{n}_{i}$ and $\bm{x}^{n}_{j}$ is regarded as an outlier. Benefit from the known 3D skeleton, we can inference its location in 3D space for correlation measurement. As shown in Figure~\ref{fig:Joints_Filter}, We design a greedy method to remove it by comparing them to $\widehat{\mathbf{X}}^n_{t}$, which is estimated by ${\mathbf{X}}^n_{t}$ and motion model (\eg, linear motion, 3D Kalman Filter~\cite{5958966}.) Afterward, the back-project-ray $\mathbf{r}$ of joint $\bm{x}^n_{i}$ can be obtained from the following formula:
\begin{equation}
    \label{eq:back-project-ray}
    \mathbf{r}^n_{i,c} = (\mathbf{K}\mathbf{R})_{c}^{+} \cdot \bm{x}^{n}_{i,c} + \mathbf{o_c}
\end{equation}
where $\mathbf{K},\mathbf{R} \in \mathbb{R}^{3\times3}$ are intrinsic and rotation matrix, respectively, and $+$ stands for the inverse matrix. $\mathbf{o}_c$ is the center of camera $c$ with respect to global coordinate system. We then compute the 3D point-to-line distance between $\widehat{\mathbf{X}}^n_{t}$ and $\mathbf{r}^n_i$. At last, we remove the joint with larger distance as the outlier of 3D skeleton to get a robust reconstruction.

% -------------------------------------------------------------------------------------------------------------
\noindent\textbf{3D Pose Reconstruction.} In our method, we use Direct Linear Transformation (DLT) for triangulation. In some cases, only one 2D pose is associated to the 3D skeleton yet triangulation needs at least two points to process. To avoid this problem, a set $\mathbb{P}$ of 2D poses is collected from a small range of time $\tau$. However, traditional triangulation is based on the condition that all the 2D points across views are from the same time, a penalty is set for the time interval, which is referenced from~\cite{chen2020cross}. But sometimes, all the joints are wrong predicted, often happen at lower arms and legs, we regard it as a missing joint $\mathbf{X}_{t, miss}^{n}$. Here we compensate $\mathbf{X}_{t, miss}^{n}$ with $\widehat{\mathbf{X}}_{t}^{n}$. Though \textit{joints filter} is adopted, slight noisy estimation still remains. We conduct temporal information to smooth the 3D pose trajectory with Gaussian filter for further enhancing the performance.

% -------------------------------------------------------------------------------------------------------------
\noindent\textbf{Initialization of 3D pose.} 
After association process in Section~\ref{sec:partial_matching}, each extracted 2D pose is labeled as either matched or unmatched. For those unmatched 2D poses from different views possibly forms a new 3D skeleton. Hence, to initialize a new 3D pose, we utilize epipolar constraint to associate unmatched 2D poses across views. Assuming that there is a set $\mathbb{U}_c$ of unmatched 2D poses in camera $c$, we measure the correlation of unmatched 2D poses from other views and $\mathbb{U}_c$ by Eq.~\ref{eq:epipolar_distance}. The final affinity matrix $\mathbf{E}=\sum^{N}_{n=1} \mathbf{E}^n$ can also be solved by Hungarian algorithm~\cite{Kuhn55thehungarian}. The process is iterative, meaning that we handle cameras one by one. After each association, there might still remain unmatched poses $\mathbb{U}'_c$ that aren't captured in camera $c$ but captured in others. We then add $\mathbb{U}'_c$ into $\mathbb{U}_c$. While completing association of all views, the \textit{joints filter} is also applied to get a robust reconstruction for initial 3D skeleton with a slight difference. Since there is no former 3D skeleton to reference, we have to solve it another way. The epipolar matrix of the $n^{\text{th}}$ joint $\mathbf{E}^n$ is still required, and an outlier exist while $\mathbf{E}^n_{i,j}$ is a negative value. We then compute the $\sum_{j=0} \mathbf{E}^n_{i,j}$ of $\bm{x}^{n}_{i}$ and $\bm{x}^{n}_{j}$, and remove the joint with smaller sum, which has a weaker affinity with others. Finally, the initial 3D pose can be reconstructed via Sec~\ref{sec:3D_Reconstruction}. 
\begin{algorithm}[t]
   \caption{Estimation and Tracking procedure}
   \label{alg:estimation_and_tracking}
   \SetAlgoLined
   \DontPrintSemicolon
   \SetNoFillComment
   \footnotesize
   \KwIn{\\
   Cameras $c \in \mathbf{C}$\\
   New 2D poses $\bm{x}_{t} \in \mathbf{p}$ \\
   Previously tracked 3D skeletons $\mathbf{X}_{t'} \in \mathbf{P}$ at time $t'$}
   \KwOut{\\
   Tracked 3D skeletons with new 3D poses $\mathbf{X}_{t} \in \mathbf{P}$ at time $t$
   }
   
   Initialization: $\mathbf{P} \leftarrow \emptyset$; 
   $\mathbf{A} \in \mathbb{R}^{|\mathbf{P}| \times |\mathbf{p}|}$ \;

   \tcc{cross-view association}
   \ForEach{$c \in \mathbf{C}$}{
       \ForEach{$i,\ \mathbf{X}_{t'} \in \mathbf{P}$ }{
          \ForEach{$j,\ \bm{x}_{t,c} \in \mathbf{p}$}{
             $\mathbf{A}(i,j) \leftarrow \operatorname{PartAwareAssociation}(\bm{x}_{t,c},\mathbf{X}_{t'})$ \;
          }
       }
       $\textit{Match}(\mathbf{X}_{t'},\bm{x}_{t,c}) \leftarrow \operatorname{Hungarian Algorithm}(\mathbf{A})$ \;
   }
    
    $\tau \leftarrow TimeInterval$\\
   \tcc{update 3D skeleton}
   \ForEach{$\mathbf{X}_{t'} \in \mathbf{P}$}{
      $\mathbb{P} \leftarrow \{\bm{x}_{t'_{c},c} | c\in \mathbf{C}; 0\leq t-t'_{c} < \tau\}$\;
      $\hat{\mathbf{X}}_t \leftarrow \operatorname{CurrentPoseEstimation}(\mathbf{X}_{t'})$\;
      $\hat{\mathbb{P}} \leftarrow \operatorname{JointsFilter}(\mathbb{P},\hat{\mathbf{X}}_t)$\;
      $\mathbf{X}_t \leftarrow \operatorname{3DReconstruction}(\hat{\mathbb{P}},\hat{\mathbf{X}}_t)$\;
   }

%   \tcc{target initialization}
%   $\mathbb{U} \leftarrow unmatched\ 2D\ poses\ of\ Camera\ 1$\\
%   \ForEach{$c \in \{\mathbf{C} - c_1\}$}{
%      $\mathbb{U}_c \leftarrow unmatched\ 2D\ poses\ of\ Camera\ c$\;
%      $\mathbf{E} \in \mathbb{R}^{|\mathbb{U}| \times |\mathbb{U}_c|}$\;
%      \ForEach{$\bm{x}_t \in \mathbb{U}$}{
%       \ForEach{$\bm{x}_{t,c} \in \mathbb{U}_c$}{
%           $\mathbf{E} \leftarrow \operatorname{Epipolar Constraint}(\mathbb{U}, \mathbb{U}_c)$ \;
%           $\textit{Match}(\bm{x}_t,\bm{x}_{t,c}),\mathbb{U}'_c \leftarrow \operatorname{Hungarian Algorithm}(\mathbf{E})$\;
%           $\mathbb{U} \leftarrow \mathbb{U} \cup \mathbb{U}'_c$
%       }
%      }
%   }
   
%   \ForEach{$\bm{x}_\textit{cluster} \in \mathbb{U}$}{
%       \If{$\operatorname{Length}(\bm{x}_\textit{cluster}) \ge 2$}{
%          $\hat{\mathbb{P}} \leftarrow \operatorname{JointsFilter(\bm{x}_\textit{cluster})}$\;
%          $\mathbf{X}_t \leftarrow \operatorname{3D Reconstruction}(\hat{\mathbb{P}})$ \;
%          $\mathbf{P} \leftarrow \mathbf{P} \cup \{ \mathbf{X}_t \}$ \;
%       }
%   }
\end{algorithm}
%------------------------------------------------------------------------
\begin{table*}[t]
\resizebox{1\textwidth}{!}{
\begin{tabular}{|c|c|c|c|c|c|c|c|c|c|}
\hline
\multicolumn{10}{|c|}{\thead{Campus Dataset, PCP(\%)}}                                                  \\ \hline
        & Belagiannis & Ershadi-N & Bridgeman & Tanke & Dong & Chen & Tu         & Zhang &               \\
Method  & \etal~\cite{belagiannis20153d} & \etal~\cite{ershadi2018multiple} & \etal~\cite{Bridgeman_2019_CVPR_Workshops} & \etal~\cite{tanke2019iterative} & \etal~\cite{dong2019fast} & \etal~\cite{chen2020cross} & \etal~\cite{voxelpose} & \etal~\cite{20204DAssociation} & \textbf{Ours} \\ \hline
Actor1  & 93.5        & 94.2      & 91.8     & 98    & 97.6 & 97.1 & 97.6       & -     & 98.37          \\
Actor2  & 75.7        & 92.9      & 92.7     & 91    & 93.3 & 94.1 & 93.8       & -     & 93.76          \\
Actor3  & 85.4        & 84.6      & 93.2     & 92.2  & 98.0 & 98.6 & 98.8       & -     & 98.26          \\ \hline
Average & 84.5        & 90.6      & 92.6     & 95.7  & 96.3 & 96.6 & \textcolor{blue}{96.7} & -     & \textcolor{red}{96.79} \\ \hline
\multicolumn{10}{c}{}                                                                                   \\
\hline
\multicolumn{10}{|c|}{\thead{Shelf Dataset, PCP(\%)}}                                                   \\ \hline
        & Belagiannis & Ershadi-N & Bridgeman & Tanke & Dong & Chen & Tu         & Zhang &               \\
Method  & \etal~\cite{belagiannis20153d} & \etal~\cite{ershadi2018multiple} & \etal~\cite{Bridgeman_2019_CVPR_Workshops} & \etal~\cite{tanke2019iterative} & \etal~\cite{dong2019fast} & \etal~\cite{chen2020cross} & \etal~\cite{voxelpose} & \etal~\cite{20204DAssociation} & \textbf{Ours} \\ \hline
Actor1  & 75.3        & 93.3      & 99.7     & 99.8  & 98.8 & 99.6 & 99.3       & 99.0  & 99.14          \\
Actor2  & 69.7        & 75.9      & 92.8     & 90    & 94.1 & 93.2 & 94.1       & 96.2  & 95.41         \\
Actor3  & 87.6        & 94.8      & 97.7     & 98    & 97.8 & 97.5 & 97.6       & 97.6  & 97.64         \\ \hline
Average & 77.5        & 88.0      & 96.7     & 95.9  & 96.9 & 96.8 & 97.0 & \textcolor{red}{97.6} & \textcolor{blue}{97.39} \\ \hline
\end{tabular}
}
\caption{Quantitative comparison on the Campus and Shelf dataset. The results of other methods are taken from respective papers. ``\textcolor{red}{red}" and ``\textcolor{blue}{blue}" represent the first and second highest scores, respectively.}
\label{tbl_campus_shelf}
\end{table*}

\begin{table}[t]
\resizebox{0.47\textwidth}{!}{
\begin{tabular}{r|cccc}
\hline
\multicolumn{1}{r|}{}          & \multicolumn{4}{c}{PCP(\%)}                           \\ \cline{2-5} 
\multicolumn{1}{r|}{Campus}                & Ua    & La    & Ul    & Ul              \\ \hline
Tanke \etal~\cite{tanke2019iterative}     & 97.7  & 84.3  & 99.3   & 99.0           \\ 
Chen \etal~\cite{chen2020cross}           & 98.6  & \textbf{84.6}  & -     & -               \\
$^\dagger$Chen \etal~\cite{chen2020cross}      & 99.8  & 84.2  & 100.0 & 100.0                                   \\
\textbf{ours}                                       & \textbf{99.8}  & 84.3  & \textbf{100.0} & \textbf{100.0}  \\ \hline
\multicolumn{1}{r|}{Shelf}                 & Ua    & La    & Ul    & Ll              \\ \hline
Tanke \etal~\cite{tanke2019iterative}     & 98.0  & 86.3  & 100.0 & 100.0           \\ 
Chen \etal~\cite{chen2020cross}           & 98.7  & 87.7  & -     & -               \\
$^\dagger$Chen \etal~\cite{chen2020cross}      & 98.8  & 89.9  & 100.0 & 99.9                                   \\
\textbf{ours}                                       & \textbf{98.8}  & \textbf{90.5}  & \textbf{100.0} & \textbf{100.0}  \\ \hline
\end{tabular}
}
\caption{PCP scores of human parts on different approaches. ``U, L, a, l" represent upper, lower, arms and legs. $^\dagger$ stands for our re-implementation of~\cite{chen2020cross} with same 2D pose estimation method as ours.}
\vspace{-0.3cm}
\label{tbl:PCP_score_of_human_parts}
\end{table}
% -------------------------------
\section{Experimental Results} 
\label{sec:exp}
\subsection{Benchmark Datasets}
\textbf{Campus dataset~\cite{belagiannis20153d}.} A small dataset includes three actors and two pedestrians walking in an outdoor environment and captured by three calibrated cameras. There are 2000 frames in the video. The $350\sim470^{\text{th}}$ and $650\sim750^{\text{th}}$ frames were used for evaluation; others were used for training. It also provides an evaluation metric, percentage of correct parts (PCP), to quantify the accuracy of detected 3D keypoints on six parts of the human body, \ie head, torso, upper arm, lower arm, upper leg, and lower leg. We noted that most of the actors in standard evaluation set videos are well captured and insufficient to test the present approach hardly. Therefore, the additional $471\sim649^{\text{th}}$ frames were added into the evaluation set. These frames are more demanding while more people were walking around.
\\
\textbf{Shelf dataset~\cite{belagiannis20153d}.} In comparison with Campus, it's a more complex dataset captured by five calibrated cameras in an indoor scenario where up to four people were dissembling a shelf at the same time. As a result, people will be shaded by the shelf or each other frequently in the video and increase the difficulty of 3D pose reconstruction. The video contains 3200 frames in total. The $300\sim600^{\text{th}}$ frames were used for evaluation; others were used for training. We follow the same evaluation protocol as the Campus dataset to quantify the accuracy of detected 3D keypoints.
\\
\textbf{CMU Panoptic~\cite{joo2017panoptic}.} In contrast to Campus and Shelf, the Panoptic dataset is a large dataset released by CMU Perceptual Computing Lab. They built an enclosed studio with 480 VGA cameras and 31 HD cameras installed on it. The videos contain multi-person 
engaging in several social activities. For the lack of the groundtruth of 3D poses, only qualitative video and image results are presented on this dataset. The qualitative results are shown in supplementary.

\begin{table}[t]
\centering
\begin{tabular}{|c|c|c|c|}
\hline
\multicolumn{4}{|c|}{\thead{Campus Extended Testing Set, PCP(\%)}}                                                  \\ \hline
        & Dong & $^\dagger$Chen &               \\
Method  & \etal~\cite{dong2019fast} & \etal~\cite{chen2020cross} & \textbf{Ours} \\ \hline
Actor1  & 97.55        & 98.37       & 98.37               \\
Actor2  & 93.33        & 95.31       & 95.65               \\
Actor3  & 98.19        & 98.13       & 98.41               \\ \hline
Average & 96.36        & 97.27       & \textcolor{red}{97.47}               \\ \hline
\end{tabular}
\caption{Quantitative comparison on the Campus extended testing set. The results of other methods are taken from respective papers. ``\textcolor{red}{red}" indicates the highest scores.}
\label{tbl_campus_estended}
\end{table}

\subsection{Comparison with other approaches}
We compare our proposed method with the following approaches. Belagiannis \etal~\cite{belagiannis20153d} develop a 3DPS-based model with temporal consistency thereafter become a baseline on Campus and Shelf datasets. Ershadi-N. \etal~\cite{ershadi2018multiple} develop a clustering algorithm to reduce the 3D state space by optimally clustering 3D candidate joints. Bridgeman \etal~\cite{Bridgeman_2019_CVPR_Workshops} develop a 2D pose association algorithm with error correction to match 2D poses in multiply views more robustly after constructing 3D poses. Tanke \etal~\cite{tanke2019iterative} develop an iterative greedy matching to remove outliers and fill-in missing joints by temporal 3D pose averaging. Dong \etal~\cite{dong2019fast} develop a multi-way matching with circle consistency to match 2D poses in multiple views optimally after constructing 3D poses. Chen \etal~\cite{chen2020cross} develop an efficient algorithm by leveraging temporal consistency to find out 2D correspondences across views. An incremental reconstruction that can deal with unsynchronized streaming video and fast-moving limbs is also designed. Tu \etal~\cite{voxelpose} develop a cuboid proposal network to localize 3D poses by integrating multi-view projected 2D pose heatmaps. Zhang \etal~\cite{20204DAssociation} develop a 4D association graph to calculate relevance between 2D and 3D joints across views and frames to obtain the best matching connection when constructing 3D poses.

Table~\ref{tbl_campus_shelf} demonstrates the PCP score of our proposed method and the above approaches evaluated on Campus and Shelf datasets. We reach state-of-the-art on Campus and competitive score on Shelf. In particular, our approach significantly improves Actor1's results on Campus , and as for Actor2 in Shelf, who suffering from severe occlusion, our method can better tackle the condition than most previous works. A further comparison was demonstrated in Table~\ref{tbl:PCP_score_of_human_parts}. We compared four human parts to prove that our proposed reconstruction method is valid. Lower arms that have larger motion amplitudes often suffer from motion blur or occlusion. In most works, the PCP score of lower arms performs under 90\%; however, our proposal successfully reach the highest accuracy up to 90.5\% on Shelf. 

Furthermore, we also evaluate some approaches~\cite{dong2019fast,chen2020cross} on the extended Campus testing set. With three actors and a pedestrian walking around not only enhance the difficulties of associations but also increase the frequency of partial occlusions, Dong \etal~\cite{dong2019fast} method slightly degrade the performance(96.36\%). On the other hand, ours and Chen \etal~\cite{chen2020cross}, which is our re-implementation, maintain considerable results on the extended frames(97.47\% and 97.27\% respectively). We conjecture that with the help of part-aware measurement and \textit{joints filter}, our approach achieves 0.2\% improvement. The result is shown in Table~\ref{tbl_campus_estended}. Thus, we can confirm that our approach can effectively and robustly construct 3D pose of people, who suffer from severe occlusion. 

\subsection{Ablation Studies}
\label{sec:ablation_study}
In this section, we give some ablation studies to verify that each component is beneficial to our approach. We compare to the baseline with body-aware measurement in association and without \textit{joints filter}s.
\begin{figure}[t]
  \centering
  \includegraphics[width=1\linewidth]{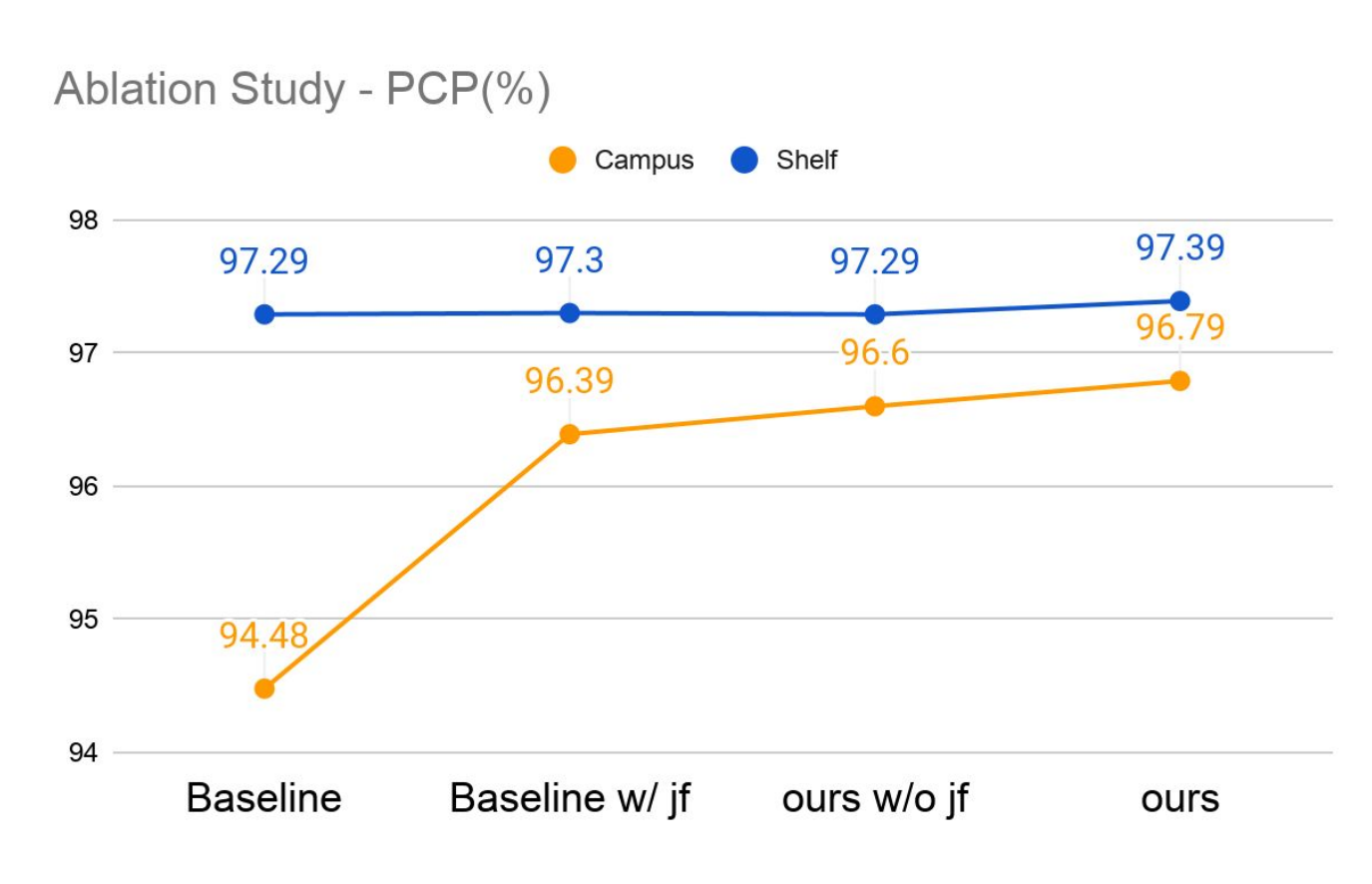}
  \caption{Ablation study on Campus and Shelf datasets. ``jf" means \textit{joints filter}, and ``ours" indicates that part-aware measurement and \textit{joints filter} are employed. As shown, applying either part-aware measurement or \textit{joints filter} performs better and by using both of them further improved the PCP result.}
  \label{fig:ablation_study}
\end{figure}

\begin{figure}[t]
  \centering
  \includegraphics[width=1\linewidth]{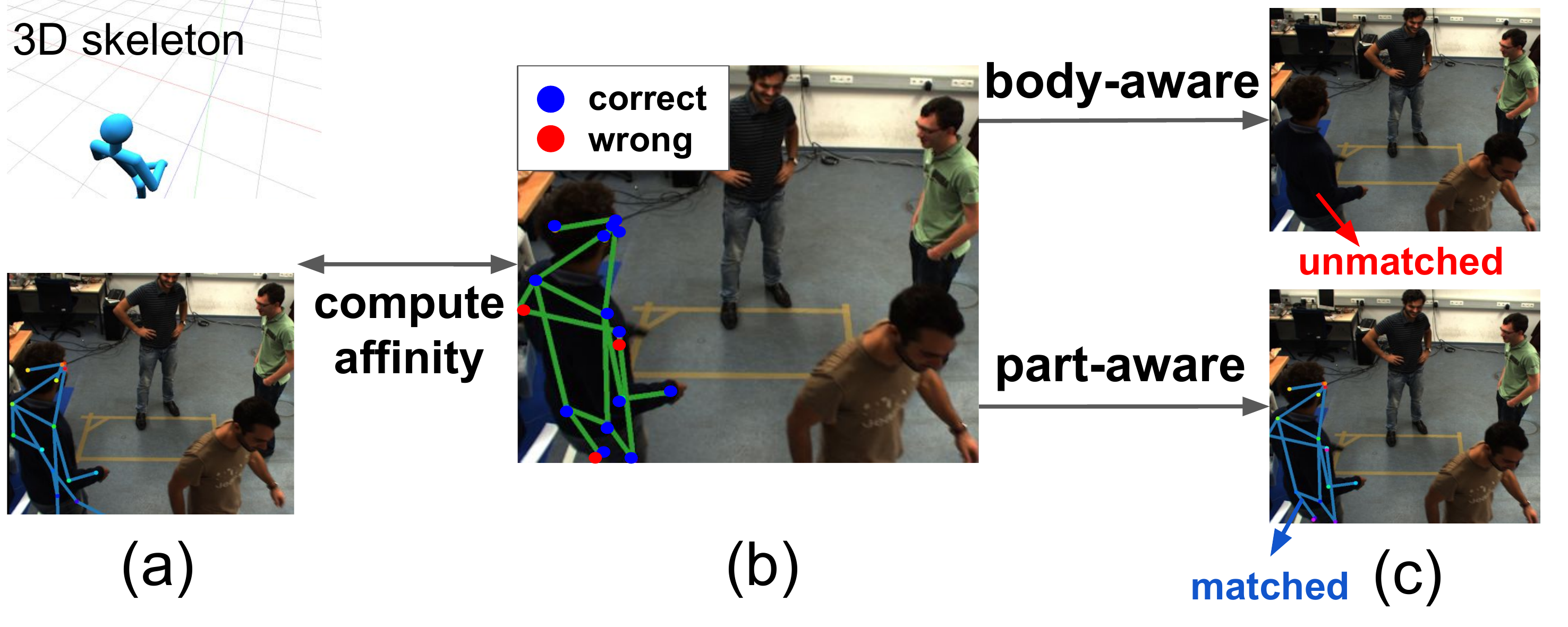}
  \caption{Visual comparison of body-aware and part-ware association on Shelf dataset. (a) a 3D skeleton and its re-projection (b) the corresponding estimated 2D pose. Due to the out-of-view of lower body, the enormous 2D estimation error of ankles leads to weak ankle affinities and affects the affinity of entire body to mis-match in body-aware association. (c) Therefore, we adopt part-aware association by neglecting the few wrong predictions but only comparing affinities of others to reach a robust matching result.
  }
  \label{fig:match_in_body_or_parts}
\end{figure}
\begin{figure*}[t]
    \centering
    \includegraphics[width=1\linewidth,scale=0.5]{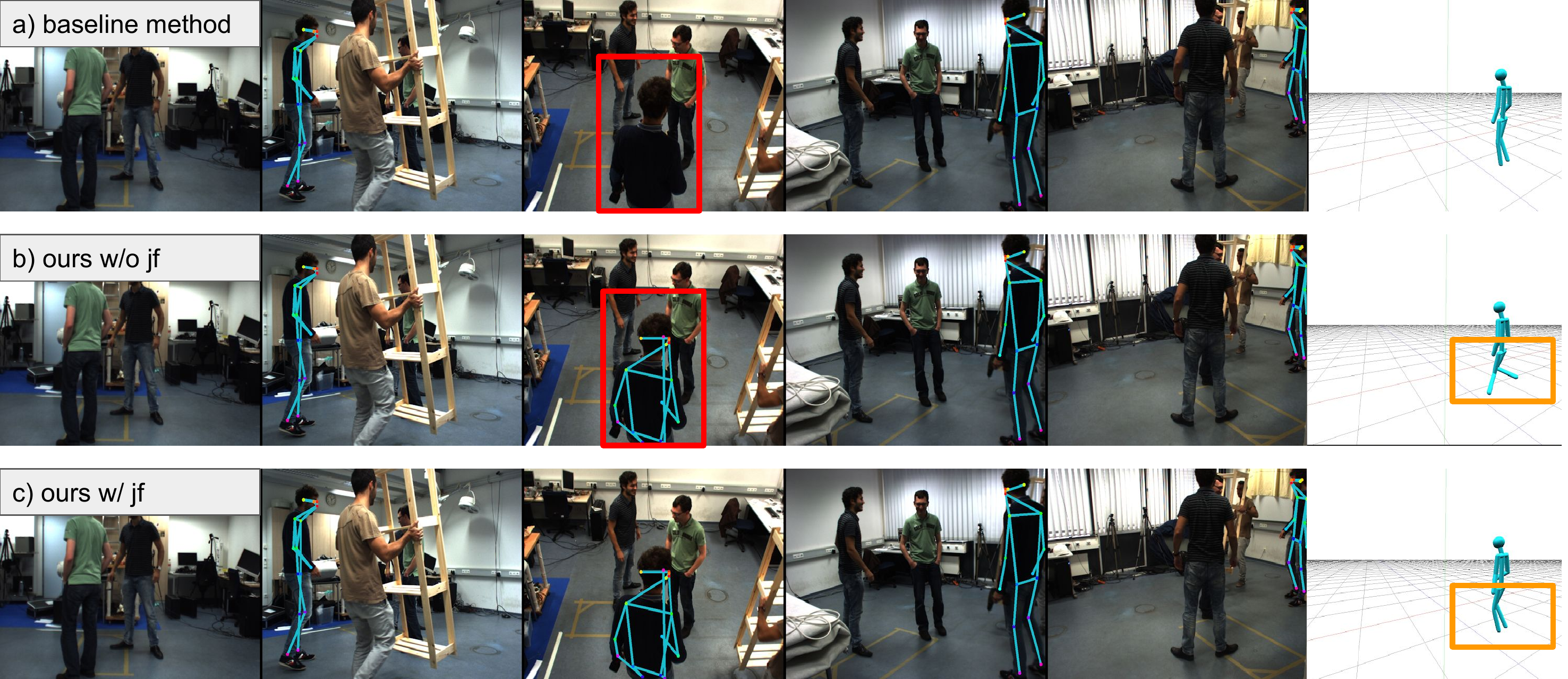}
	\caption{
	A visualization of ablation studies. From left to right indicate the camera's order, and the last column is the 3D skeleton from $c_3$'s viewpoint. As shown in \textcolor{red}{red} bounding boxes, the black-shirt man's lower body is out of $c3$'s FOV, leading to wrong prediction.	(a) body-aware association failed to associate the 2D pose in $c3$ to the corresponding 3D skeleton. (b) altering association with part-aware measurement succeeds. Yet due to the outliers of his lower body, the legs of 3D skeleton perform weird poses (\textcolor{orange}{orange} box.) (c) further improves the 3D pose reconstruction of (b) with joint filter.
	}
	\label{fig:joint_filter_visualization}
\end{figure*}

\noindent\textbf{Measure in body or part-aware?} To verify that measuring in part-aware is effective, we only change the method building associations. The other modules remain the same as the baseline. In some cases, even though most joints have good predictions, there were still seldom joints with an enormous error that significantly affect the affinity and causing unmatched consequence(an example is given in Figure~\ref{fig:match_in_body_or_parts}). As demonstrated in Table~\ref{fig:ablation_study}, singly applying part-aware measurement enhanced approximately 2.1\% on Campus, which is a large improvement. We investigate that measuring in part-aware leads to fewer unmatched results, representing that for each 3D skeleton, there are more matched 2D poses across views. Hence, with more cross-view 2D correspondences, a better 3D skeleton can be constructed.
\\
\textbf{Effectiveness of Joints Filter?} As described in Section~\ref{sec:3D_Reconstruction}, we regard \textit{joints filter} as an important process to handle noisy points that will degrade the reconstruction accuracy. For comparison, we select the torso, lower arms to verify the impact of the \textit{joints filter}. Torso performs well (100\%) in every method due to the small range of motion. For the lower arms, which are considered the most demanding limb among whole body parts, the \textit{joints filter} method improves about 0.5$\sim$1.0\% and PCP score. Some examples on Shelf are shown in Figure~\ref{fig:joint_filter_visualization}. Additionally, to justify that \textit{joints filter} not only works in our proposal, we add it to the re-implementation of Chen \etal~\cite{chen2020cross}. We can improve the PCP score even from 97.28\% to 97.43\%, which is a 0.15\% enhancement.

\begin{table}[t]
\resizebox{0.47\textwidth}{!}{
\begin{tabular}{|r|cccc|c|}
\hline
            & \multicolumn{4}{c|}{Processing Speed on Shelf (ms)} & \multicolumn{1}{l|}{} \\ \cline{2-5}
Method      & Affi.        & Asso.       & Recon.       & Init.       & fps                   \\ \hline
Bridgeman \etal~\cite{Bridgeman_2019_CVPR_Workshops}   & -           & -          & -           & -          & 110  \\
Dong \etal~\cite{dong2019fast}        & 25          & 20         & 60          & -          & 9.5                   \\ 
Chen \etal~\cite{chen2020cross}        & -           & -          & -           & -          & \textbf{325}                   \\ 
\textbf{ours w/o jf} & \multicolumn{2}{c}{1}    & 3.5         & 1          & 181                   \\ 
\textbf{ours}        & \multicolumn{2}{c}{1}    & 8          & 1          & 100                  \\ \hline
\end{tabular}
}
\caption{Runtime comparison on Shelf. Stages `Affi.', `Asso.', `Recon.' and `Init.' stand for affinity computation, association, reconstruction and 3D skeleton initialization, respectively.}
\label{tbl:running_speed}
\end{table}
\subsection{Processing time}
In this section, we report the runtime of our algorithm on different datasets. Considering 2D pose estimation is beyond the scope, only tracking and reconstruction time are demonstrated.
\\
\textbf{Shelf.} For the pose estimation, our model takes about 600ms to produce all the 2D poses. Since most of the multi-view 3D human pose estimation methods focus on the reconstruction process, we compare the result with other's proposals on this part. Dong \etal~\cite{dong2019fast} test their experiment on GeForce 1080Ti GPU, which spent an average of 25ms to compute affinities and 20 ms for finding the cross-view association, and 60 ms for reconstruction. In contrast, our implementation runs on Intel CPU 3.70GHz. It takes about 2 ms for finding the 2D-3D association, 8 ms for reconstruction and 1 ms for the cross-view initialization.
\\
\textbf{Campus} contains fewer people and cameras, the processing speed is faster than the Shelf. It takes about 1 ms for association, 5 ms for reconstruction, and 0.5 ms for initialization.
%------------------------------------------------------------------------
\section{Conclusion}
\label{sec:conclusion}
We propose an efficient and robust multi-person 3D pose estimation and tracking from multiple views by directly associating 2D poses estimation to a 3D skeleton in each view and updating 3D poses with a part-aware approach. Our proposed part-aware association is able to achieve better matching in contrast to common body-aware methods. Moreover, joint filter also plays an important role in dealing with noisy joint estimation to get a more accurate reconstruction. In experiments, our approach achieved a competitive result on benchmark datasets without any additional training. And the ablation study verified our part-aware method is effective in partial occlusion.
\\
\\
\noindent\textbf{Acknowledgement.} This work was partially supported by Ministry of Science and Technology in Taiwan (MOST 110-2634-F-002-04).

\section{Supplementary Material}
\subsection{Implementation Details}
\textbf{Paremeters Selection.} In this work we have several parameters: $\alpha_{2D}$, $\alpha_{epi}$, $\tau$, $\varepsilon$ are thresholds of 2D velocity, epipolar distance, time interval and number of positive affinities respectively, $\lambda_{\alpha}$ is a penalty rate of time interval. Here we show our empirical selections of parameters for each dataset in Table~\ref{tbl:parameters}.  $\alpha_{2D}$, $\alpha_{epi}$, $\tau$, $\varepsilon$ are based on the frame size and the distance between people and camera. For Campus~\cite{belagiannis20153d}, the image size is $360 \times 288$ and the actors are far from cameras. Therefore, $\alpha_{2D}$ and $\alpha_{epi}$ is set to be smaller number. Yet $\varepsilon$ adjust with a strict value since actors are mostly captured completely, and we expect that 2D poses can all be accurately estimated. On the other hand, Shelf~\cite{belagiannis20153d} and Panoptic~\cite{joo2017panoptic} have larger image size and humans are captured in a small area which is close to cameras. Thus, occlusion and out-of-view often occur. We then define the parameter with a more flexible value. For other two parameters, $\tau$ and $\lambda_{\alpha}$ basically depends on the fps of video, \eg the three datates are all captured at 25 fps.
\begin{table}[h]
\centering
\begin{tabular}{|r|ccc|}
\hline
Dataset            & Campus & Shelf & Panoptic \\ \hline
$\alpha_{2D}$      & 30     & 70    & 60       \\
$\alpha_{epi}$     & 15     & 60    & 30       \\
$\tau$             & 3      & 3     & 3        \\
$\varepsilon$      & 14     & 10    & 10       \\
$\lambda_{\alpha}$ & 3      & 3     & 3        \\ \hline
\end{tabular}
\caption{Parameters selection for different datasets.}
\label{tbl:parameters}
\end{table}
\\
\noindent\textbf{Initialization Procedure.} To introduce our 3D pose initialization procedure more clearly, it is detailed in Algorithm~\ref{alg:initialization}.
\begin{algorithm}[t]
   \caption{Initialization Procedure}
   \label{alg:initialization}
   \SetAlgoLined
   \DontPrintSemicolon
   \SetNoFillComment
   \footnotesize
   \KwIn{\\
   Unmatched 2D poses ${\mathbb{U}_c | c \in \mathbf{C}}$ \\
   Previously tracked 3D skeletons $\mathbf{X}_{t'} \in \mathbf{P}$ at time $t'$}
   \KwOut{\\
   New tracked skeletons $\{\mathbf{X}_{t}\}$ at time $t$
   }
   
   Initialization: $\mathbb{U} \leftarrow unmatched\ 2D\ poses\ of\ Camera\ 1$\\
  \ForEach{$c \in \{\mathbf{C} - c_1\}$}{
     $\mathbb{U}_c \leftarrow unmatched\ 2D\ poses\ of\ Camera\ c$\;
     $\mathbf{E} \in \mathbb{R}^{|\mathbb{U}| \times |\mathbb{U}_c|}$\;
     \ForEach{$\bm{x}_t \in \mathbb{U}$}{
      \ForEach{$\bm{x}_{t,c} \in \mathbb{U}_c$}{
          $\mathbf{E} \leftarrow \operatorname{Epipolar Constraint}(\mathbb{U}, \mathbb{U}_c)$ \;
          $\textit{Match}(\bm{x}_t,\bm{x}_{t,c}),\mathbb{U}'_c \leftarrow \operatorname{Hungarian Algorithm}(\mathbf{E})$\;
          $\mathbb{U} \leftarrow \mathbb{U} \cup \mathbb{U}'_c$
      }
     }
  }
   
  \ForEach{$\bm{x}_\textit{cluster} \in \mathbb{U}$}{
      \If{$\operatorname{Length}(\bm{x}_\textit{cluster}) \ge 2$}{
         $\hat{\mathbb{P}} \leftarrow \operatorname{JointsFilter}(\bm{x}_\textit{cluster})$\;
         $\mathbf{X}_t \leftarrow \operatorname{3D Reconstruction}(\hat{\mathbb{P}})$ \;
         $\mathbf{P} \leftarrow \mathbf{P} \cup \{ \mathbf{X}_t \}$ \;
      }
  }
\end{algorithm}

\subsection{Qualitative Results}
\begin{figure*}[t]
    \centering
    \includegraphics[width=1\linewidth]{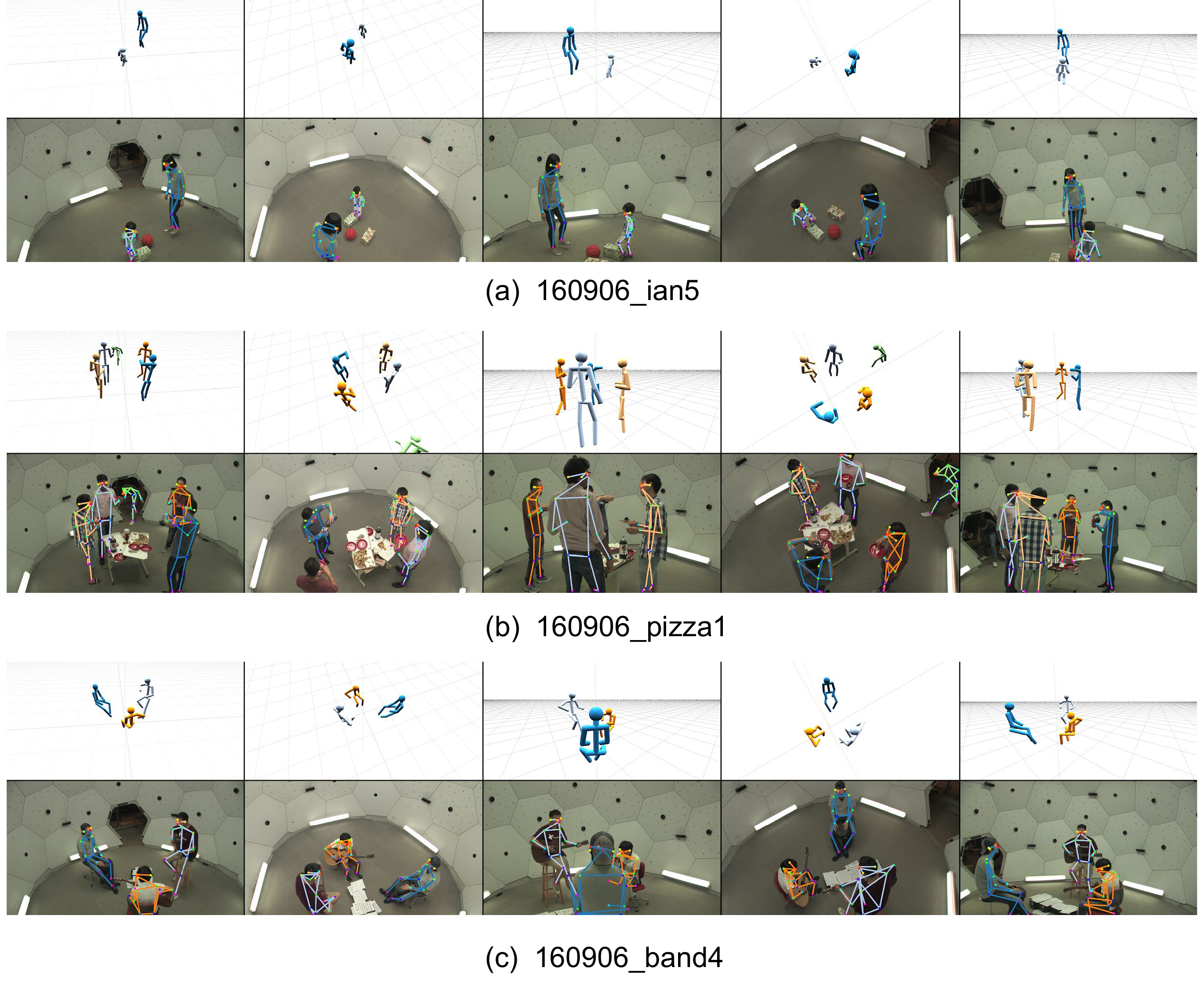}
    \caption{Qualitative results of Panoptic, three sub-datasets is demonstrated.}
    \label{fig:Panoptic}
\end{figure*}

\begin{figure}[t]
    \centering
    \includegraphics[width=1\linewidth]{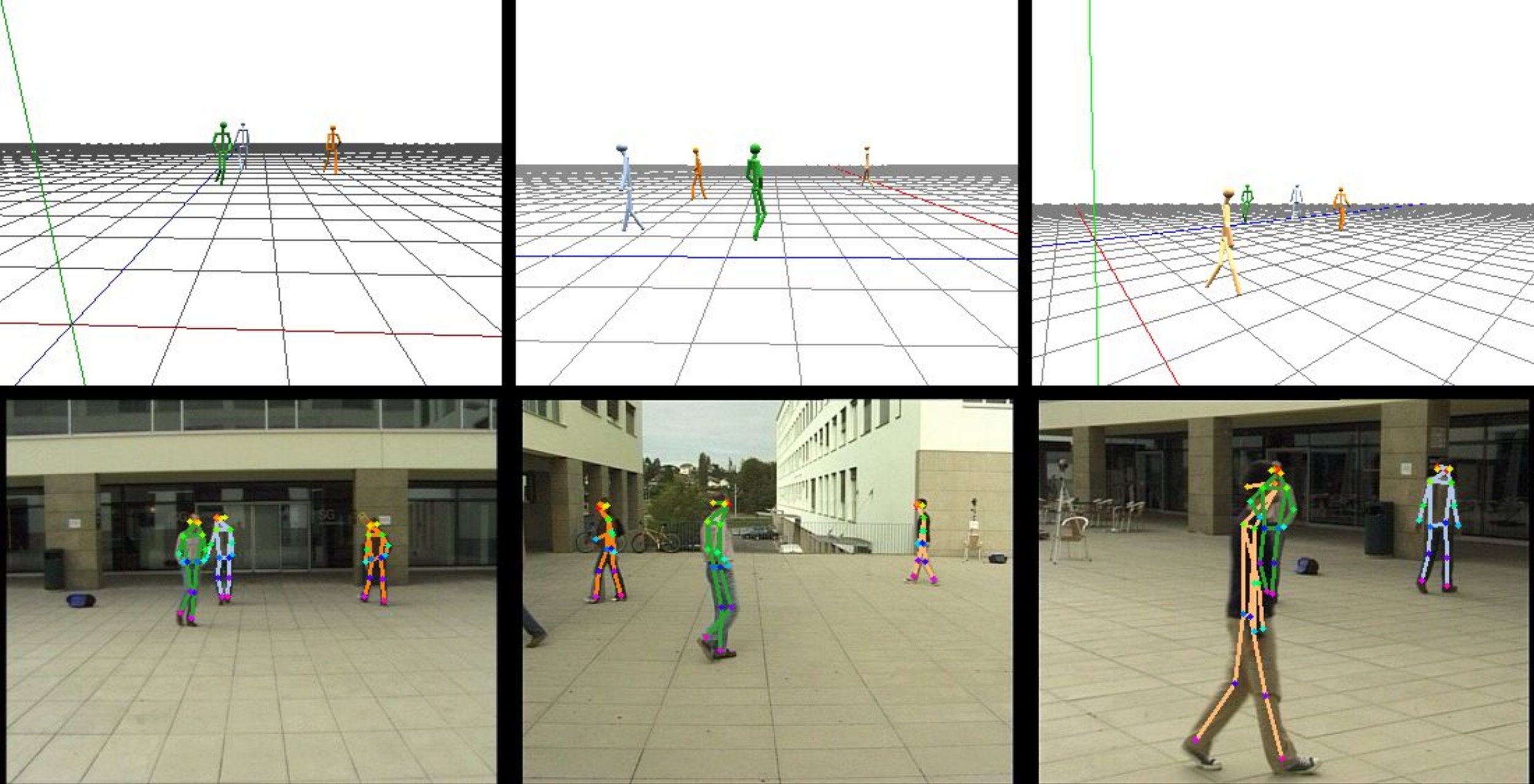}
    \caption{Qualitative result of Campus}
    \label{fig:Campus}
\end{figure}

\begin{figure}[t]
    \centering
    \includegraphics[width=1\linewidth]{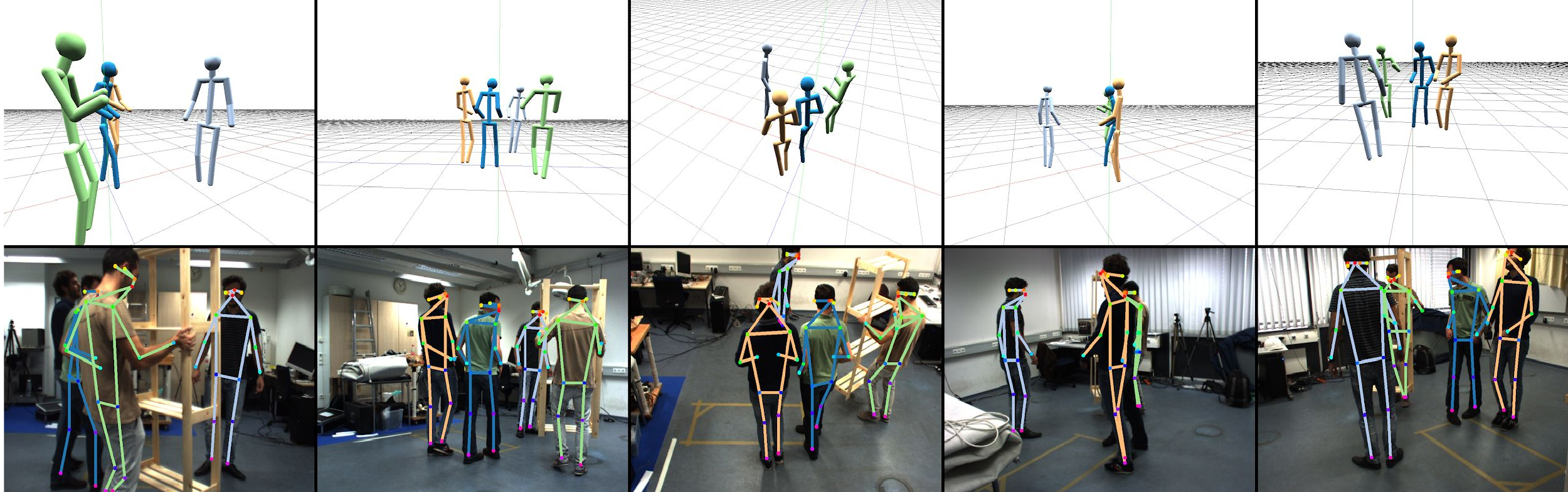}
    \caption{Qualitative result of Shelf}
    \label{fig:Shelf}
\end{figure}

Here, we demonstrate more qualitative results of our approach on three datasets in Figure~\ref{fig:Panoptic}, Figure~\ref{fig:Campus} and Figure~\ref{fig:Shelf}.

{\small
\bibliographystyle{ieee_fullname}
\bibliography{egbib}
}

\end{document}